# Integrated BIM and Machine Learning System for Circularity Prediction of Construction Demolition Waste

Abdullahi Saka[1]; Ridwan Taiwo[2]; Nurudeen Saka[3]; Benjamin Oluleye[2]; Jamiu Dauda[4]; Lukman Akanbi[5*]

[1] *Associate Professor in Sustainable Digital Construction and Enterprise; Westminster Business School, University of Westminster, London, UK.*

[2] *Researcher; Department of Building and Real Estate, the Hong Kong Polytechnic University, Hong Kong.*

[3] *Researcher; School of Built Environment, Engineering and Computing, Leeds Beckett University, Leeds, UK.*

[3] *Senior Lecturer; School of Built Environment, Engineering and Computing, Leeds Beckett University, Leeds, UK.*

[5] *Professor of Applied AI and Big Data Analytics; Birmingham City Business School, Faculty of Business, Law and Social Sciences, Birmingham City University, Birmingham, UK*

*Corresponding Author: lukman.akanbi@bcu.ac.uk

# Integrated BIM and Machine Learning System for Circularity Prediction of Construction Demolition Waste

**Abstract**

Effective management of construction and demolition waste (C&DW) is crucial for sustainable development, as the industry accounts for 40% of the waste generated globally. The effectiveness of the C&DW management relies on the proper quantification of C&DW to be generated. Despite demolition activities having larger contributions to C&DW generation, extant studies have focused on construction waste. The few extant studies on demolition are often from the regional level perspective and provide no circularity insights. Thus, this study advances demolition quantification via Variable Modelling (VM) with Machine Learning (ML). The demolition dataset of 2280 projects were leveraged for the ML modelling, with XGBoost model emerging as the best (based on the Copeland algorithm), achieving $R^2$ of 0.9977 and a Mean Absolute Error of 5.0910 on the testing dataset. Through the integration of the ML model with Building Information Modelling (BIM), the study developed a system for predicting quantities of recyclable and landfill materials from building demolitions. This provides detailed insights into the circularity of demolition waste and facilitates better planning and management. The SHapley Additive exPlanations (SHAP) method highlighted the implications of the features for demolition waste circularity. The study contributes to empirical studies on pre-demolition auditing at the project level and provides practical tools for implementation. Its findings would benefit stakeholders in driving a circular economy in the industry.

## 1.0 Background

The construction industry is one of the most vital sectors as it provides infrastructural support for other industries and has significant economic impacts (Pheng & Hou, 2019). It contributes about 13% of the world's GDP and employs more than 7% of the global population (Eurostat, 2023). However, it also has detrimental impacts on the environment. For instance, the building sector is responsible for 38% of global energy-related $CO_2$ emissions, 35% of global energy use, and about 55% of global electricity use (UNEP, 2022). Similarly, the construction industry is the largest consumer of natural resources – using 40% of global raw material - and, as such, the largest source of waste generation – producing 40% of global waste (Malia *et al.*, 2013; Oluleye *et al.*, 2022). These highlighted the unsustainability trend of the current practices in the industry. However, the trend seems unabated due to the prevailing linear economy, rapid urbanization, demand for infrastructure, and increased global population (Saka *et al.*, 2019).

The waste generated in the construction industry are categorized into construction waste (CW), renovation waste (RW), and demolition waste (DW) with construction and demolition waste (C&DW) waste accounting for the largest percentage. In the UK, 137.8 million tonnes of C&D waste were generated in 2018, accounting for 62% of waste generated in the UK (Defra, 2022). The waste management hierarchy recommends prevention, reuse, recycling, recovery, and disposal in order of effectiveness (Ajayi & Oyedele, 2017). However, despite strategies such as designing out waste via optimization and using standard materials, among others, it is impossible to prevent waste generation in construction sites (Saka *et al.*, 2019). Thus, the focus has been on 'zero avoidable' waste in the construction industry, emphasising the lower strategies in the hierarchy – reuse, recycling, recovery, and disposal (Osmani, 2011). This has also been supported with the push for a circular economy in the industry as against the linear approach – 'take, make, dispose' (Bilal *et al.*, 2020). The effectiveness of these lower management strategies for construction and demolition waste depends on the availability of adequate information about the likely waste to be generated, such as material type and quantities (Akanbi *et al.*, 2020). As such, waste quantification and estimation (WQ&E) are key requirements for circularity in the industry. These would enable efficient planning in advance for C&D waste management and inform key decisions such as the size of waste containers and frequency of replacement (Malia *et al.*, 2013). Also, WQ&E enhances the optimization of activities such as deconstruction, decommissioning, and demolition by providing necessary information for stakeholders (Akanbi *et al.*, 2018).

Consequently, extant studies have focused on waste estimation and quantification for effective management of construction and demolition waste in the industry. Wu *et al.* (2014) highlighted that these studies could be regional or project level waste quantification. The study categorised the quantification methods into site visit (SV), generation rate calculation (GRC) method, lifetime analysis (LA) method, classification system accumulation (CSA) method and variables modelling (VA) method. SV entails visiting the site for direct or indirect measurement of C&DW; and GRC involves leveraging waste generation rates such as waste generated/tonnes/person/year, or waste generated/area of project (Wu *et al.*, 2014). On the other hand, LA method assumes that the total volume of material will be the eventual demolition waste; CSA provides waste generation per specific unit such as material or component, and VA entails modelling waste prediction based on key features (Lam *et al.*, 2019). Kartam *et al.* (2004) employed SV via analysis of trucks arriving at a landfill to quantify regional level construction and waste demolition in Kuwait. de Guzman Baez *et al.* (2012) adopted VA and proposed two equations for C&DW in railway construction in Spain based on features such as kilometers of railway, viaduct, tunnels, and numbers of junctions, drainage & underpasses. In Portugal, Bernardo *et al.* (2016) used a VM with a dataset of 50 demolition projects to propose a waste quantification model at the regional level with predictors such as population density, age of the building, building density, and type of land occupation. Li and Zhang (2013) employed GRC and CSA to develop a web-based system for quantifying building construction waste. Similarly, Cheng and Ma (2013) leveraged GRC and CSA to developed a BIM-based system for quantifying renovation and demolition wastes in Hong Kong. Lam *et al.* (2019) employed GRC and CSA for quantification of construction waste generation in new builds in Hong Kong. SV + GRC + CSA were implemented by Yu *et al.* (2019) via Google imagery (indirect measurement), waste generation rate (WGR) and gross floor area (GFA) for quantification of regional level waste during urban renewal in Shenzhen. Kamma and Jha (2022) adopted GRC to compute C&DW waste generated at regional level using permit data in India.

Despite the fact that demolition activities have made a greater contribution to C&DW, previous studies have mainly focused on construction waste (Yuan & Shen, 2011). Extant studies on demolition waste quantification and estimation – pre-demolition audit – often leverage a manual approach to measure waste on site, which is time-consuming (Akanbi *et al.*, 2020). While other studies have adopted quantification methods such as LA, GRC, and CSA, which could provide information about demolition without a site visit, these methods do not provide circularity options. For instance, these studies revealed an estimate for generating demolition waste but provided no insights about what quantity to reuse, recycle, or dispose of. As such, practical applications of these extant studies are limited for efficient pre-demolition auditing. In addition, leveraging methods such as variable modelling (VM), which could model demolition waste generation quantities based on relevant predictors, 'has not got a wide application' despite its capability to forecast C&DW due to lack of datasets (Kamma *et al.*, 2022; Wu *et al.*, 2014).

Over the years, with the availability of more datasets, studies have been employing machine learning (ML) algorithms for VM of C&DW quantification. At the regional level, Lu *et al.* (2021) employed population, GDP per capital, total construction output, and floor space of newly started buildings and completed as predictors for estimating waste generation in Greater Bay Area China using Multiple Linear Regression (MLR), Decision Tree, Grey Model and Artificial Neural Network (ANN) with varying $R^2$ in the range 0.756 - 0.977. Coskuner *et al.* (2021) employed ANN to predict construction and demolition waste in Bahrain using input variables of population, GDP, construction records & land area with an $R^2$ of 0.91. Islam *et al.* (2019) adopted regression analysis to model construction material generation in Bangladesh with $R^2$ ranging from 0.73 – 0.96. On the other hand, at the project level, Cha *et al.* (2023) used 160 demolition datasets to model waste generation rates using location, usage, structure, wall type, and roof type as predictors in South Korea. K-nearest neighbour, linear regression, principal component analysis (PCA), and decision tree algorithm were employed with hybrid PCA-K-nearest neighbour as the best-performing model with $R^2$ of 0.897. Cha, Moon*, et al.* (2022) employed a demolition dataset of 784 buildings to model demolition waste prediction in China using ANN, Support Vector Machine Regression (SVMR), and categorical principal component analysis (CATPCA) with the region, use, structure type, wall material type, roofing material, and gross floor area as input variables. Hybrid CATPCA-SVMR was the best performing with an $R^2$ of 0.594. MLR was also adopted by Soultanidis and Voudrias (2023) to model demolition waste generation of construction materials using 45 residential building demolition datasets in Greece with height, areas, levels, and openings as input variables, while Teixeira *et al.* (2020) employed MLR with 18 data points for waste quantification in Brazil with $R^2$ of 0.81.

Consequently, there has been a growing application of ML modelling for VM of C&DW prediction over the years. However, all these extant studies do not provide circularity insights for the demolition waste and adopt a single output modelling, which hinders effective demolition auditing and waste management planning. As such, this current study aims to develop a BIM-based ML system to predict building demolition waste circularity. This will predict DW quantities that could be recycled, reused, and disposed of. The following are the specific objectives of the study:

i) To model demolition waste generation using multi-output machine learning algorithms

ii) To optimise and assess the best-performing machine learning model for circularity prediction of demolition waste

iii) To develop a BIM-based ML system for circularity prediction of demolition waste

This study contributes to the growing body of knowledge on demolition waste quantification at the project level using the VM approach. It provides an easy-to-use BIM integrated system for circularity prediction of demolition waste that supports pre-demolition auditing. The developed system would enable stakeholders to forecast and plan for waste to be reused, recycled, or disposed of during building project demolition. The rest of the paper is structured as follows: The second section presents the methodological steps employed in achieving the aim of the study. Section 3 provides the results of the analysis and development, while the discussion of the findings is presented in Section 4. The conclusions and implications of the study are presented in Section 5.

## 2.0 Methodology

Three sequential phases are employed to achieve the aim of the study, as shown in Figure 1.

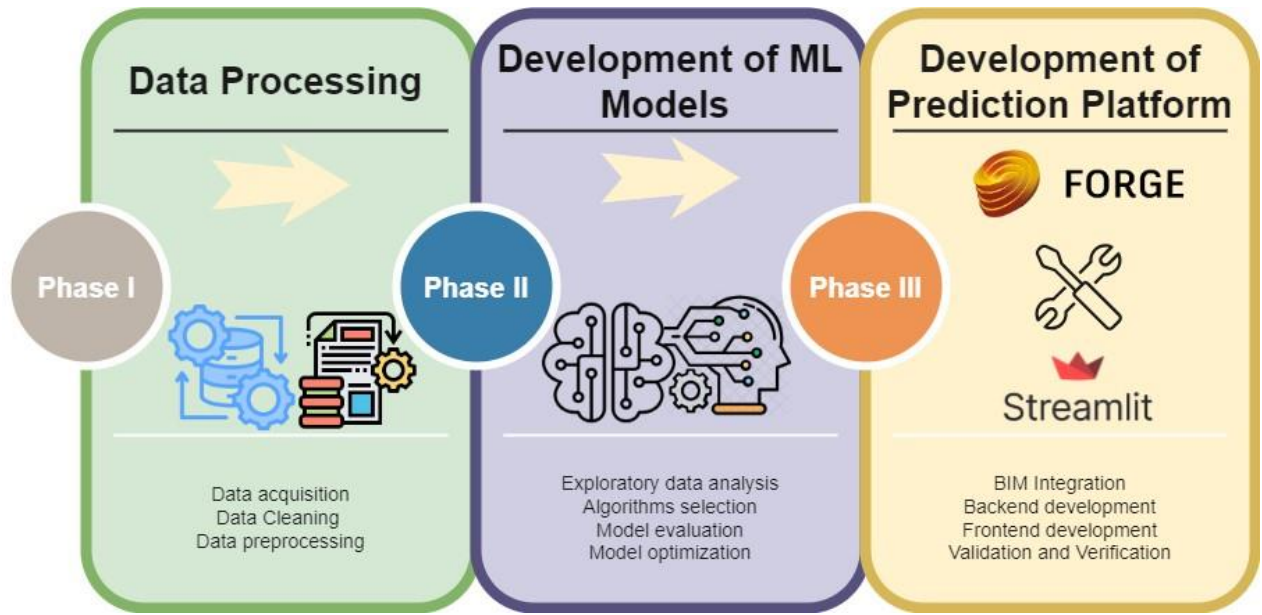

Figure 1: Research Methods

### 2.1 Data processing (Phase I)

The demolition dataset was obtained from the UK members of Institution of Demolition Engineers (IDE) and the National Federation of Demolition Contractors (NFDC), which has also been used in previous studies. This dataset was used because it is difficult to obtain a comprehensive dataset for demolition due to current practices in the industry. Obtaining quantities of demolition waste has proved to be near impossible due to limited resources (space and facilities) for sorting demolition debris on-site. Thus, the common practice involves transporting the debris to nearby Waste Transfer Stations (WTS). A more thorough segregation process is conducted at these stations, sorting the materials for various end-uses like recycling, reuse, or landfilling. However, these sorted debris are not mapped to actual buildings from which the wastes were obtained (Akanbi *et al.*, 2020). Consequently, demolition records of 2280 projects in the UK with details of the materials, frame type, usage type, gross floor area, volume, no of levels, demolition quantity, recycle quantity, and reuse quantity were retrieved for this study. Predictors of demolition waste were selected based on extant studies within the limit of available variables in the dataset (Cha *et al.*, 2023; Soultanidis *et al.*, 2023; Teixeira *et al.*, 2020). The descriptive analysis of the selected variables (dataset) is shown in Table 1.

Table 1: Descriptive Analysis of Demolition Waste Dataset from 2280 UK buildings.

| Factor | Description | Mean | Standard deviation | Minimum value | Maximum value |
|---|---|---|---|---|---|
| GFA | Gross Floor Area, representing the total floor space of the building | 1962.75 | 2684.70 | 4.80 | 10051.00 |
| Volume | Total volume of the building, reflecting its overall size | 6910.82 | 11350.36 | 9.60 | 76380.00 |
| Number of levels | The total number of floors or levels in the building | 2.00 | 1.07 | 1.00 | 7.00 |
| Recyclable material | Quantity of demolition materials that can be recycled | 500.07 | 792.89 | 0.98 | 4453.63 |
| Reusable materials | Amount of demolition materials that can be reused | 108.87 | 303.03 | 0.11 | 1936.34 |
| Landfill materials | Amount of demolition materials destined for landfill | 74.76 | 206.39 | 0.00 | 1229.50 |

## 2.2 Development of ML Models (Phase II)

### 2.2.1 Predictive Algorithms

Predictive Modelling deals with applying 'statistical models or data mining algorithms to data for the purpose of predicting new or future observations' (Shmueli, 2010). Machine learning (ML) algorithms that include Decision Tree, K-Nearest Neighbours, Random Forest, eXtreme Gradient Boosting (XGBoost), and LightGBM were selected for the modelling due to their ability to handle large datasets, complex relationships, and adaptation for multivariate outputs (Egwim *et al.*, 2021; Shehadeh *et al.*, 2021). Also, these algorithms have been employed for construction and demolition waste quantification in extant studies (Coskuner *et al.*, 2021; Lu *et al.*, 2021). In addition, deep learning algorithms were not employed for the data due to the simplicity and structure of the dataset.

*2.2.1.1 Decision Tree*

A Decision Tree (DT) recursively splits the data into subsets based on the value of the input features. This process creates a tree-like model of decisions, where each internal node represents a decision on an attribute, each branch represents the outcome of the decision, and each leaf node represents the predicted output (regression value). Given a dataset $D$ with $n$ samples and $m$ features, the algorithm seeks to split $D$ into subsets $D_1, D_2, ..., D_k$ based on a feature *F* that optimises a certain criterion (like minimization of variance). The process can be formally expressed as (Breiman et al., 1984):

$$D = \bigcup_{i=1}^{K} D_i \quad (1)$$

$$Var(D) = \sum_{i=1}^{k} \frac{|D_i|}{|D|} \cdot Var(D_i) \quad (2)$$

where $Var(D)$ is the variance of the target variable in the dataset, and $|D_i|$ denotes the number of points in the subset $D_i$.

The construction of a DT follows a top-down approach, commencing from the root node. This root node represents the entire dataset, encompassing all features and target values. The essence of building the tree involves splitting this root node into child nodes, thereby partitioning the dataset into increasingly more homogeneous subsets in terms of the target variable (Yussif et al., 2023). Each split is based on a specific feature that best segregates the data at that stage. A schematic representation of the DT algorithm is shown in Figure 2.

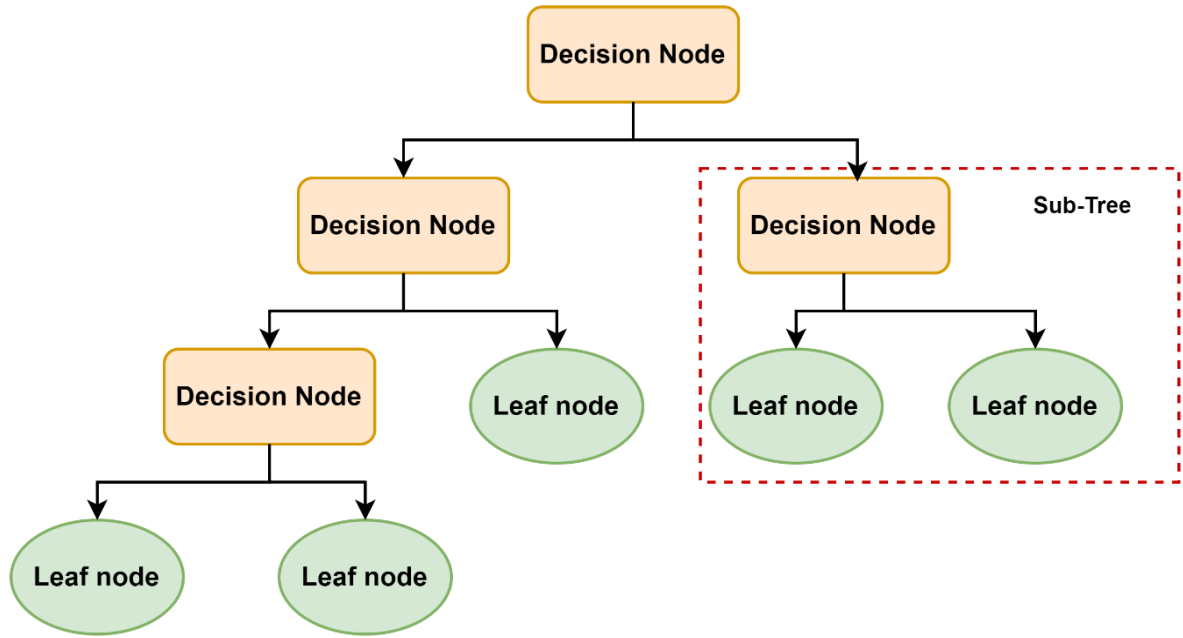

Figure 1: Schematic representation of Decision Tree

*2.2.1.2 K-Nearest Neighbours*

K-Nearest Neighbours (KNN) is a method that utilizes the concept of feature similarity to make predictions (Patrick & Fischer, 1970). In this method, the prediction for a new data point is determined by considering the outputs of the 'K' closest data points in the training set. This approach is based on the underlying assumption that similar data points will likely yield similar outputs. In applying this method to a dataset $D$ that comprises $n$ samples, each characterized by $m$ features, the KNN regression works by calculating the predicted value $\hat{y}$ for a new data point $x$. This prediction is made by averaging the target values of the K-nearest neighbors of $x$ (Wang & Ashuri, 2016).

Identifying these neighbours involves calculating the distance between the new data point and all other points in the dataset, followed by selecting the closest' K' points. The mathematical expression for this prediction is given by Equation 3 (Samet, 2008).

$$\hat{y}(x) = \frac{1}{K} \sum_{i \in N_k(x)} y_i \qquad (3)$$

where $N_k(x)$ is the set of K-nearest neighbors to the data point $x$, and $y_i$ represents the target values of these neighbors. This formula provides a quantitative means to predict the outcome for new data points based on the knowledge gleaned from the most similar instances in the training set. A schematic representation of the KNN algorithm is given in Figure 3. The figure shows that the new data point belongs to category A as three neighbours surround it compared to two neighbours from category B.

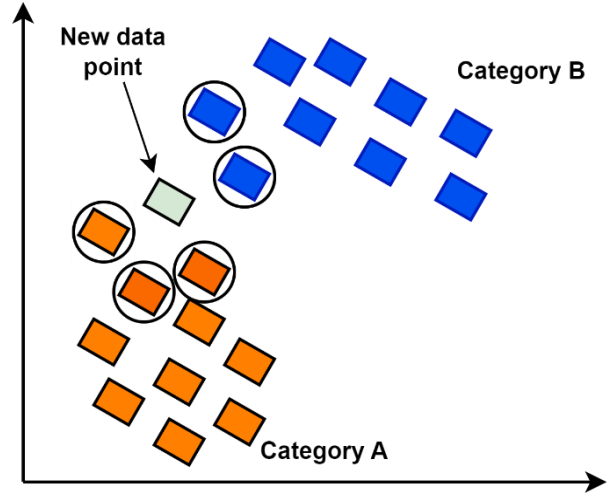

Figure 2: A schematic representation of K-Nearest Neighbours algorithm

*2.2.1.3 Random Forest*

The Random Forest (RF) algorithm creates a 'forest' of DTs using random subsets of the training data and features at each split in the construction of the trees. For regression problem in this study, it operates by building multiple DTs and then averaging their predictions to arrive at a more accurate and stable estimate. This ensemble approach leverages the strength of multiple models to improve predictive performance. Each tree in the RF is built from a different sample of the data (i.e., $D_1(x,y), ..., D_k(x,y)$) where the sampling is done with replacement, a method known as bootstrap sampling (Breiman, 2001). Moreover, when splitting each node during the construction of trees, a random subset of features is considered for the split. The number of features considered at each split is a parameter of the model and can be adjusted to optimize performance.

The output of the RF is calculated by averaging the predictions of all the individual trees on a per-data-point basis. If $Y_i$ is denoted as the prediction of the $i^{th}$ tree, and there are $N$ trees in the forest, the prediction $\hat{y}$ for a new data point $x$ is given by Equation 4 (Breiman, 2001). Figure 4 shows a schematic representation of the RF algorithm.

$$\hat{y}(x) = \frac{1}{N} \sum_{i=1}^{N} Y_i(x) \qquad (4)$$

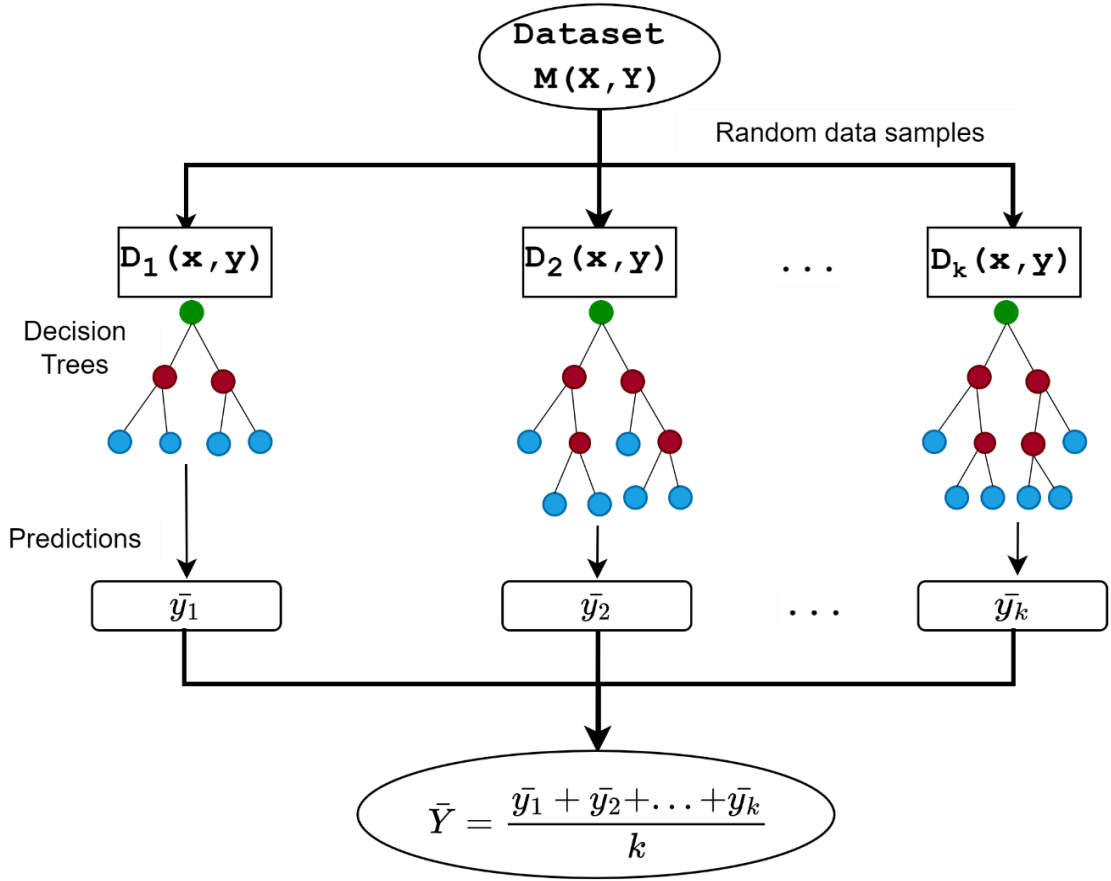

Figure 3: A schematic representation of the Random Forest algorithm

### *2.2.1.4 XGBoost*

XGBoost, extreme gradient boosting, is an ensemble learning method that builds upon the principle of boosting, where weak learners (typically decision trees) are sequentially added to correct the errors made by the preceding models. Unlike RF, which builds each tree independently, boosting methods build trees in succession, with each new tree attempting to correct the errors of the aggregate ensemble thus far (Chen & Guestrin, 2016). The XGBoost algorithm employs a number of advanced techniques to improve and speed up the boosting process. It includes a regularized model to control over-fitting, which provides it with better performance and generalization capabilities. For the regression task in this study, XGBoost predicts the target variable by learning from the data with multiple trees that sequentially focus on the residuals of prior trees.

The objective function that XGBoost optimizes consists of a loss function that measures the difference between the predicted and true values for the regression task, combined with a regularization term that penalizes the complexity of the model, as shown in Equation 5 (Ben Seghier et al., 2022; Chen & Guestrin, 2016).

$$Obj(\Theta) = L(\Theta) + \Omega(\Theta) \tag{5}$$

where $L$ is the loss function that measures the difference between the predicted value $\widehat{y_i}$ and the actual value $y_i$ and $\Omega$ represents the regularization term, which penalizes the complexity of the model, $\Theta$. It should be noted that XGBoost can handle various types of features

(numerical and categorical) and can automatically learn the non-linear interactions among them. The algorithm also has the capability to handle missing values. Figure 5 gives a schematic representation of XGBoost.

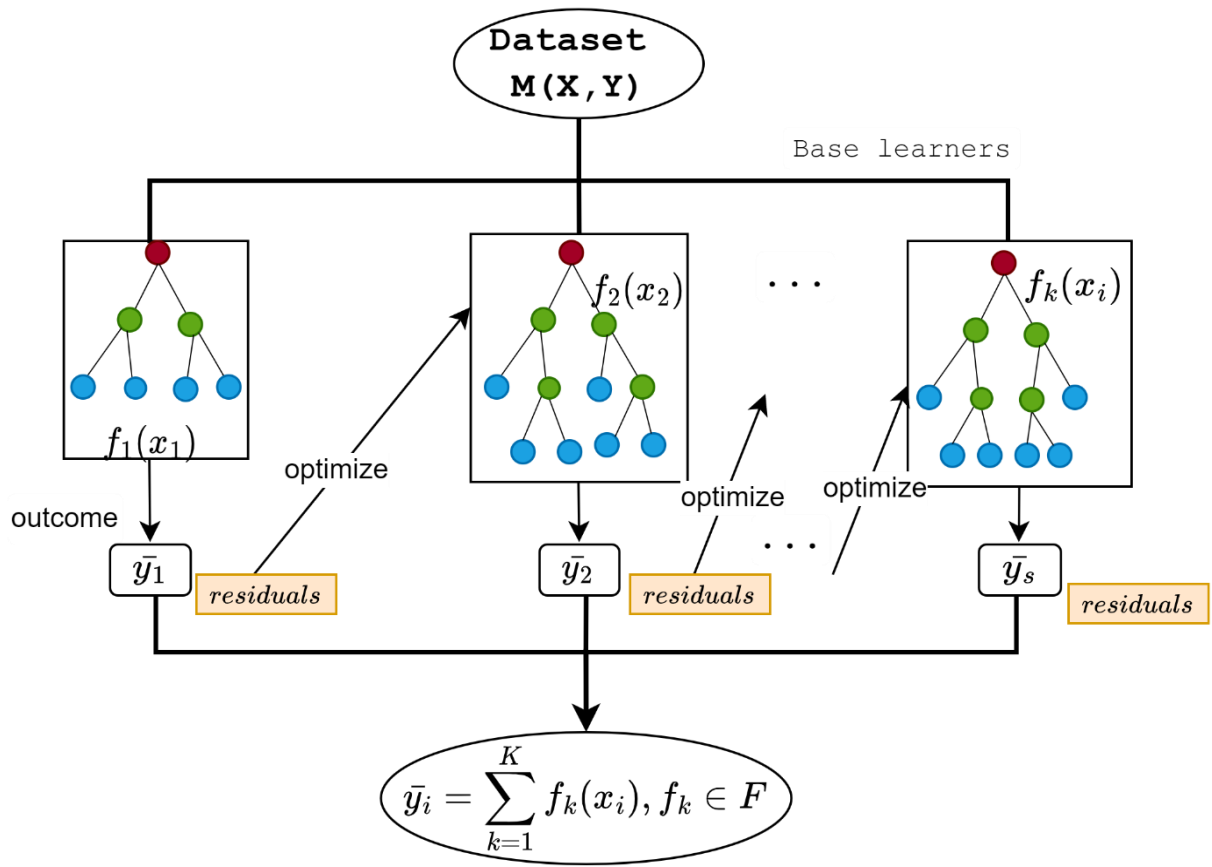

Figure 4: A schematic representation of XGBoost

### 2.2.1.5 LightGBM

Like XGBoost, LightGBM builds the model in a gradient boosting framework, but it differs by utilizing a technique called Gradient-based One-Side Sampling (GOSS) to filter out the data instances to find a split value, while still maintaining the accuracy of a given dataset (Daoud, 2019). It also uses Exclusive Feature Bundling (EFB), which reduces the number of features in a dataset by combining mutually exclusive features, thus increasing the efficiency of the algorithm. LightGBM improves the efficiency of the model by focusing on instances with larger gradients. GOSS keeps all instances with large gradients and randomly samples those with small gradients. The equation for the adjusted gradients could be represented by Equation 6 (Ke et al., 2017).

$$L'(\Theta) = \sum xi \in Al(\Theta; xi) + \frac{1}{1-\alpha}\sum xi \in Bl(\Theta; xi) \qquad (6)$$

Unlike other boosting algorithms that grow trees level-wise, LightGBM grows trees leaf-wise (see Figure 5), which can be illustrated by the following condition for selecting the leaf to grow:

$$Choose\ leaf = \underset{leaf}{argmax}\ \Delta Loss \qquad (7)$$

where $\Delta Loss$ is the loss reduction and the selected leaf is the one that maximizes this value.

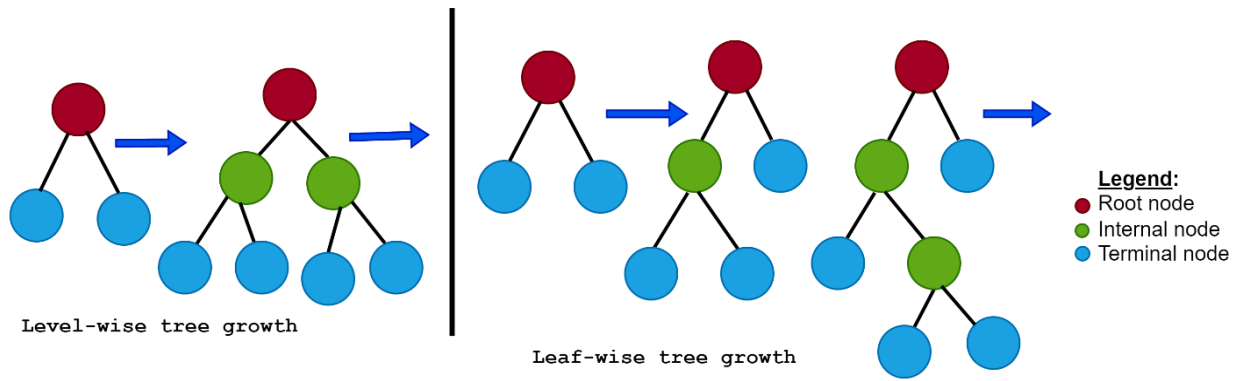

Figure 5: Level-wise and leaf-wise tree growth

## 2.2.2 Model Optimization

Bayesian Optimization (BO) is a probabilistic model-based approach for finding the minimum of a function that is expensive to evaluate. It is particularly well-suited for hyperparameter tuning, where each objective function evaluation can be very costly (Sun et al., 2021; Taiwo et al., 2023a). BO works by constructing a posterior distribution of functions (probabilistic model) that best describes the function that needs to be optimized. As the algorithm explores the space, the model is updated with the objective function's evaluations, which are used to make decisions about where in the parameter space to sample next.

The decision about where to sample next is based on a criterion that balances exploration (sampling where the model is uncertain) and exploitation (sampling where the model predicts a good objective function value). A commonly used criterion is the Expected Improvement (EI), which at any point $x$ is given by Equation 8 (Victoria & Maragatham, 2021).

$$EI(x) = E[max(0, f(x_{best}) - f(x))] \qquad (8)$$

where $f(x_{best})$ is the best objective function value observed so far, and $f(x)$ is the value of the function at new point $x$, under the model's current belief about the function.

The BO was used to optimize the hyperparameters of the five ML algorithms employed in this study. The steps followed for the optimization are outlined in Figure 7, and the hyperparameters together with their range are denoted in Table 2.

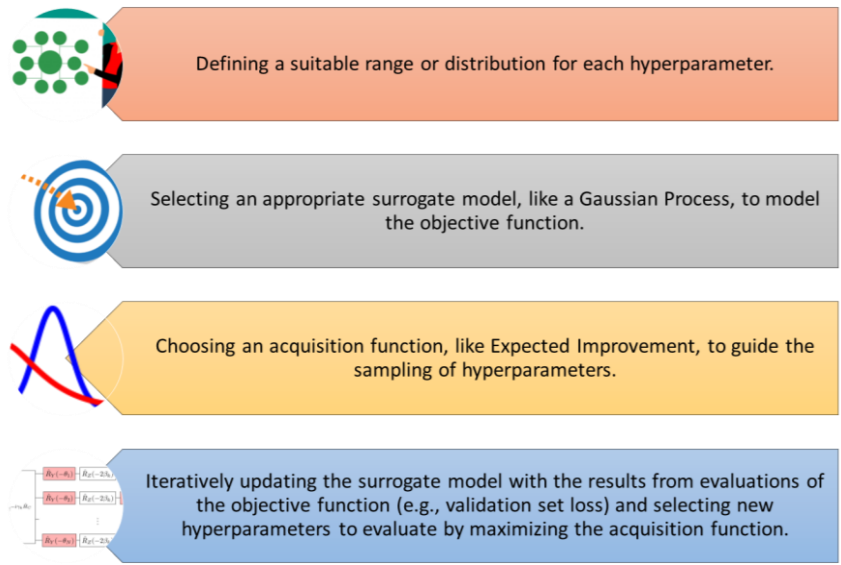

Figure 7: Hyperparameter optimization using BO

The process continues until a stopping criterion is reached, which was a set of number of iterations or no significant improvement in the objective function.

Table 1: Hyperparameter details of the models

| Models | Hyperparameters | Range |
|---|---|---|
| **DT** | Maximum depth | [1, 500] |
| | Minimum samples split | [1, 20] |
| | Minimum samples leaf | [1, 20] |
| **KNN** | Number of neighbours | [1, 10] |
| | Weights | [Uniform, Distance] |
| | Power parameter | [1, 5] |
| **RF** | Maximum depth | [1, 500] |
| | Minimum samples split | [1, 20] |
| | Minimum samples leaf | [1, 20] |
| **XGBoost** | Maximum depth | [1, 500] |
| | L1 regularization | [0, 1] |
| | L2 regularization | [0, 1] |
| | Subsample | [0.001, 1] |
| | Number of estimators | [1, 1000] |
| **LightGBM** | Maximum depth | [1, 500] |
| | L1 regularization | [0, 1] |
| | L2 regularization | [0, 1] |
| | Subsample | [0.001, 1] |

|  | Number of estimators | [1, 1000] |
| --- | --- | --- |

## 2.2.3 Model Evaluation

To quantitatively evaluate the performance of the predictive models, seven metrics are employed, which are categorized into similarity-based and dissimilarity-based metrics (see Table 3). The similarity-based metrics include the coefficient of determination ($R^2$) and Nash-Sutcliffe Efficiency (NSE) (Abyani et al., 2022). The $R^2$ metric is pivotal in understanding how well the model's predictions conform to the actual observed values. It essentially measures the proportion of variance in the observed data that the model is able to explain, thereby offering insight into the model's explanatory power. A higher $R^2$ value signifies a model that closely aligns with the observed data trends. The NSE serves a similar purpose but adds a layer of normalization. It assesses the model's predictive skill by comparing the magnitude of residual variance to the observed data variance. A higher NSE value indicates that the model's predictions are accurate and consistent with the scale of the observed data.

On the other hand, dissimilarity-based metrics, which include root mean square error (RMSE), mean absolute error (MAE), mean absolute percentage error (MAPE), similarity index (SI), and the 95% uncertainty range (U95), focus on quantifying the extent of error or deviation in the model's predictions from the actual values (Bello et al., 2023; Taiwo et al., 2023b). RMSE and MAE are the most direct measures of error magnitude, with RMSE being more sensitive to large errors due to its squaring of residuals. These metrics are essential for understanding the average magnitude and variance of errors in the model's predictions. MAE provides a more straightforward average error magnitude, making it less sensitive to outliers compared to RMSE. MAPE extends this understanding to a relative scale by expressing the error as a percentage of the actual values (Taiwo et al., 2022). The U95 metric provides an estimate of the uncertainty in the model's predictions. It essentially delineates the range within which the true values are likely to fall with a high degree of confidence. This metric is instrumental in risk assessment and decision-making processes, where understanding the reliability of predictions is as crucial as their accuracy.

Table 2: Model performance metrics

| Performance indicator | Category | Formula | Remark |
| --- | --- | --- | --- |
| R2 | Similarity-based | $1 - \frac{\sum_{i=1}^{n}(D_i - P_i)^2}{\sum_{i=1}^{n}(D_i)^2}$ $0 \leq R2 \leq 1$ | The closer the values of these metrics to 1, the better the model |
| NSE |  | $1 - \frac{\sum_{i=1}^{n}(D_i - P_i)^2}{\sum_{i=1}^{n}(D_i - \overline{D})^2}$ $-1 \leq NSE \leq 1$ |  |
| RMSE | Dissimilarity-based | $\sqrt{\frac{1}{n}\sum_{i=1}^{n}(D_i - P_i)^2}$ | The lower the values of these metrics, the better the model |
| MAE |  | $\frac{1}{n}\sum_{i=1}^{n} \lvert D_i - P_i \rvert$ |  |
| MAPE | | $\frac{1}{n}\sum_{i=1}^{n}\left|\frac{D_i - P_i}{D_i}\right|$ | |
| U95 | | $1.96\sqrt{(SD^2 - RMSE^2)}$ | |
| SI | | $\frac{RMSE}{\overline{D}}$ | |

Note: $D_i$ is the i-th average measured value of demolition materials (i.e., recycle, reuse, and landfill materials), $P_i$ is the i-th average predicted value of demolition materials, $\overline{D}$ is the arithmetic mean of the measured values of demolition materials.

### 2.2.3.1 Model ranking using the Copeland method

The Copeland method provides a systematic approach to rank the selected models based on their performance across various metrics. It involves pairwise comparisons of models across different performance metrics and aggregates these comparisons to derive an overall ranking (Furxhi et al., 2019). The procedure for applying the Copeland method in this study involves the following steps:

**i) Pairwise Comparison:** Each model is compared with every other model across all chosen performance metrics. For instance, this study has five models and six metrics; each model was compared with every other model on each metric, resulting in a series of pairwise comparisons.

**ii) Score allocation:** For each pairwise comparison, the model that performs better on a particular metric is assigned a score of +1, the model that performs poorer scores a -1, and in case of a tie, both models score a draw (assigned a score of 0).

**iii) Aggregating Scores:** The scores from each pairwise comparison are aggregated for each model. This aggregation results in a Copeland score for each model.

**iv) Wins and Loses:** A 'win' is counted for a model when it accumulates more scores in a pairwise comparison than its competitor. For instance, should Model A outperform Model B across the chosen metrics in their direct comparison, Model A is deemed to have achieved a 'win' over Model B. On the other hand, a model incurs a 'loss' when its total score is lower than that of its counterpart in a pairwise comparison.

v) **Ranking Models:** The models are then ranked based on their Copeland scores. A higher Copeland score indicates a model that consistently outperforms other models across various metrics, thus suggesting superior overall performance.

### 2.2.3.2 Model explainability using SHAP

Understanding why an ML model makes certain predictions is crucial for evaluating its trustworthiness and reliability. To provide insights into the selected ML model and explain its predictions, the SHapley Additive exPlanations (SHAP) approach is utilized (Lundberg & Lee, 2017; Taiwo, Zayed, et al., 2024). SHAP is a model interpretation technique based on game theory and Shapley values. It explains a model's predictions by calculating the contribution of each feature to the prediction. The SHAP values represent how much each feature pushed the prediction from the base value (average model output) to the current output. The SHAP values sum to the difference between the expected output of the model and the actual prediction.

Mathematically, the SHAP value φj for feature j is calculated as:

$$\emptyset_j = \sum_{S \subset N\{j\}} \frac{(|S|-1)!\cdot(|N|-|S|-1)!}{|N|!}[f(S \cup \{j\}) - f(S)] \qquad (9)$$

where *N* represents the set of all features, *S* is a subset of features excluding *j*, *f(S)* is the model output when considering features in subset *S*, and *f(S∪{j})* is the model output when feature *j* is included. Features with positive SHAP values increase the prediction, while negative values decrease it. The magnitude of the SHAP value represents the impact of that feature.

## 2.3 Development of Prediction Platform

The platform was developed using Streamlit, a Python framework for developing web applications by integrating the best-performing ML model with Autodesk Forge API to predict circularity (recycle, reuse, and landfill) of demolition waste in construction projects. The best-performing model, according to the Copeland method, is pickled and loaded to predict the demolition waste circularity of new projects. The features for the predictive model (GFA, Volume, Number of levels) were extracted from the BIM model via Autodesk Forge API, and the user input other features (frame type and usage). The platform embedded Autodesk Forge Viewer to visualize the BIM model for the proposed demolition and allow users to interact with the 3D model directly within the application. Lastly, the building parameters (from the BIM model and users) are used to make circularity predictions with the pickled ML model, and the results are displayed.

## 3.0 Analysis

This section presents the results of the predictive modelling and the system development of a demolition waste circularity predictive system.

### 3.1 Optimized ML Models

Five ML models were developed using the training dataset - KNN, DT, RF, XGBoost, and LightGBM. Hyperparameter optimization was conducted on each model using Bayesian Optimization to improve performance. The optimized models were evaluated on the training and testing dataset using dissimilarity-based (RMSE, MAE, MAPE, SI, and U95) and similarity-based performance metrics ($R^2$ and NSE). Tables 4 and 5 present the results of the optimized models using training and testing datasets, respectively. 80% of the data was used for training, while the remaining 20% was used for testing. According to Table 4, DT, RF, XGBoost, and LightGBM performed similarly on the training set. They achieved RMSE scores between 46.48 and 46.57, $R^2$ values of ~0.9959, and NSE scores of 0.9726-0.9727. However, KNN had the worst performance on the training set with the highest RMSE of 48.5206 and lowest $R^2$ of 0.9956. This indicates that it had the largest errors and least explanatory power on the training data compared to the other models.

**Table 4: Performance metrics of the optimized models using the training dataset**

| Models | RMSE | MAE | MAPE | SI | U95 | R2 | NSE |
|---|---|---|---|---|---|---|---|
| KNN + BO | 48.5206 | 4.8950 | 1.0071 | 8.7303 | 60.5861 | 0.9956 | 0.9703 |
| DT + BO | 46.4804 | 5.0503 | 0.9992 | 8.3734 | 58.1088 | 0.9959 | 0.9727 |
| RF + BO | 46.4818 | 5.0624 | 1.1533 | 8.4227 | 58.1793 | 0.9959 | 0.9727 |
| XGBoost + BO | 46.5742 | 5.0910 | 1.1076 | 8.3954 | 58.2293 | 0.9959 | 0.9726 |
| LightGBM + BO | 46.4806 | 5.0583 | 1.0760 | 8.3736 | 58.1093 | 0.9958 | 0.9727 |

A similar trend is noticed on the testing dataset (see Table 5). XGBoost and RF had the lowest RMSE of 40.3379 and 40.9294 and the highest $R^2$ of 0.9977 and 0.9976. This was a significant improvement over their training set RMSE of 46.5742 and 46.4818, showing their robustness against overfitting. KNN showed a gap between training and testing performance. Its testing RMSE of 45.5524 was much better than its training RMSE of 48.5206. Hence, the testing results highlight the importance of evaluating models on unseen data, as differences in generalization ability become apparent. Generally, it could be said that all the models are capable of predicting the amount of demolition materials, as they exhibit low error and high explanatory power (Taiwo, Yussif, et al., 2024).

**Table 5: Performance metrics of the optimized models using the testing dataset**

| Models | RMSE | MAE | MAPE | SI | U95 | R2 | NSE |
|---|---|---|---|---|---|---|---|
| KNN + BO | 45.5524 | 4.0258 | 0.9389 | 7.1288 | 56.7997 | 0.9971 | 0.9779 |
| DT + BO | 40.9294 | 3.1630 | 0.7890 | 6.4172 | 51.3077 | 0.9976 | 0.9821 |
| RF + BO | 40.9062 | 3.8202 | 0.8342 | 6.4138 | 51.1028 | 0.9976 | 0.9821 |
| XGBoost + BO | 40.3379 | 3.7874 | 0.9289 | 6.3294 | 50.4121 | 0.9977 | 0.9826 |
| LightGBM + BO | 41.416 | 3.9881 | 1.4677 | 6.4776 | 51.6962 | 0.9976 | 0.9817 |

Figures 8 - 11 depict the plots of predicted vs actual values for the amount of recyclable, reusable, and landfill materials on the training and testing datasets for each ML model. The points clustered along the diagonal line in each plot indicate the strong correlation between the predicted and actual values. Therefore, the closer the fit of the points to the diagonal, the better the model's predictive performance.

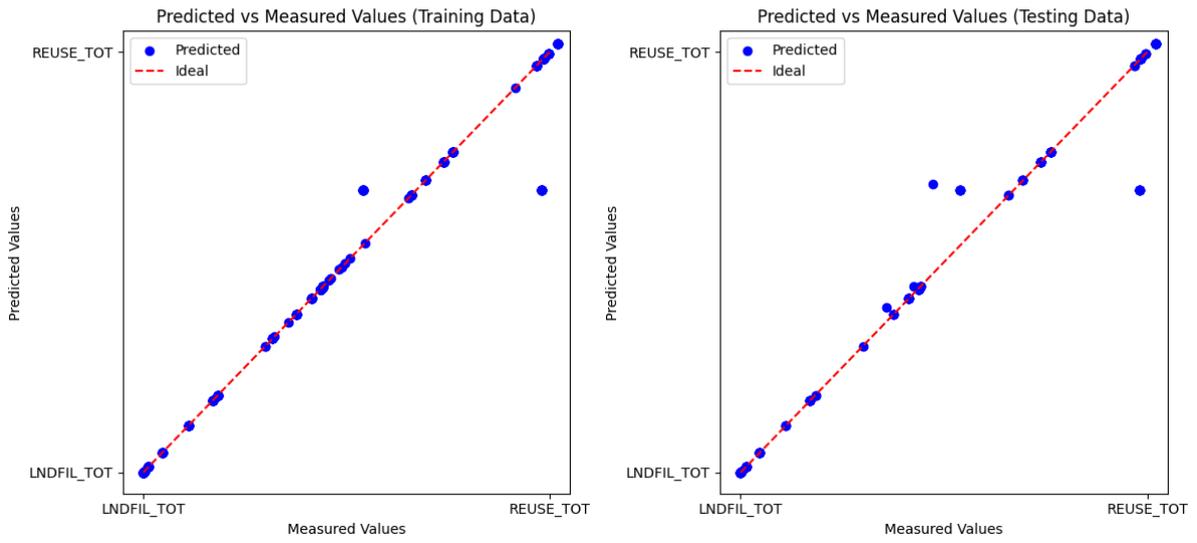

Figure 8: Actual value versus predicted value for the demolition materials – K-Nearest Neighbors +BO

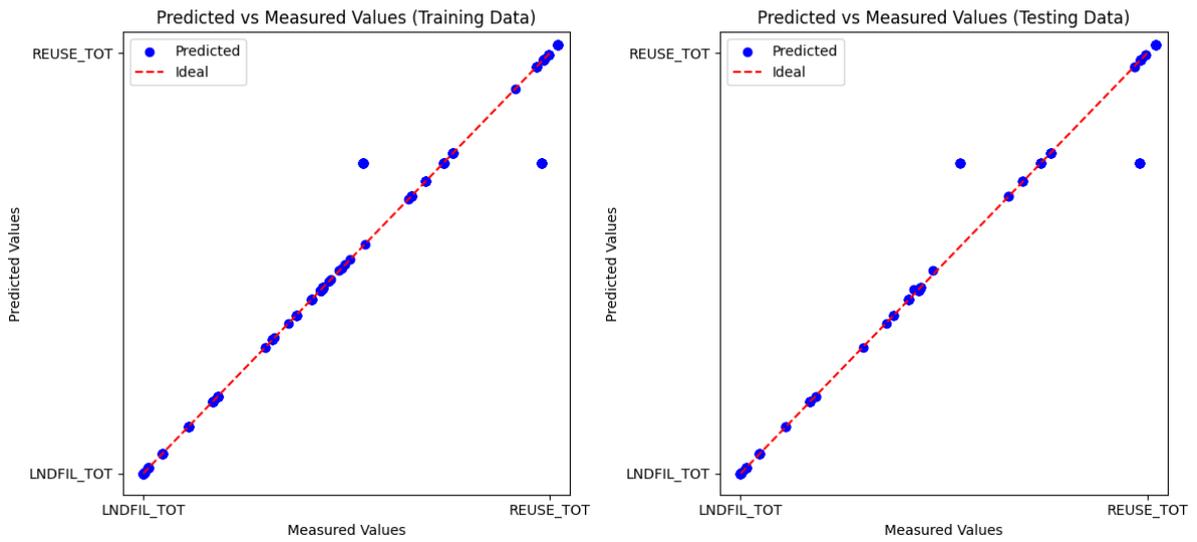

Figure 9: Actual value versus predicted value for the demolition materials – Decision Tree +BO

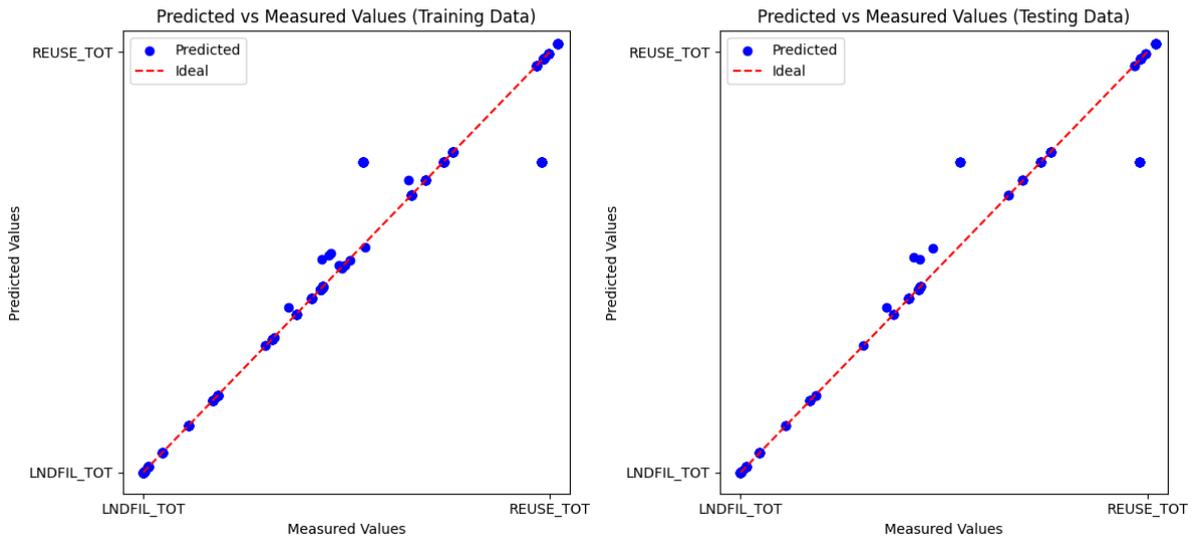

Figure 10: Actual value versus predicted value for the demolition materials – Random Forest +BO

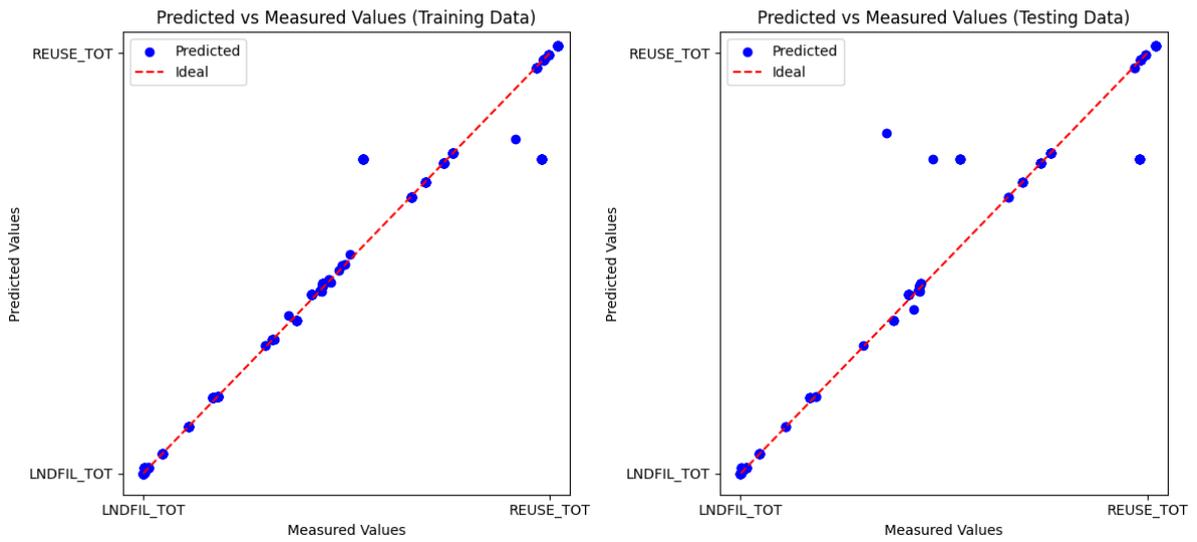

Figure 11: Actual value versus predicted value for the demolition materials – XGBoost +BO

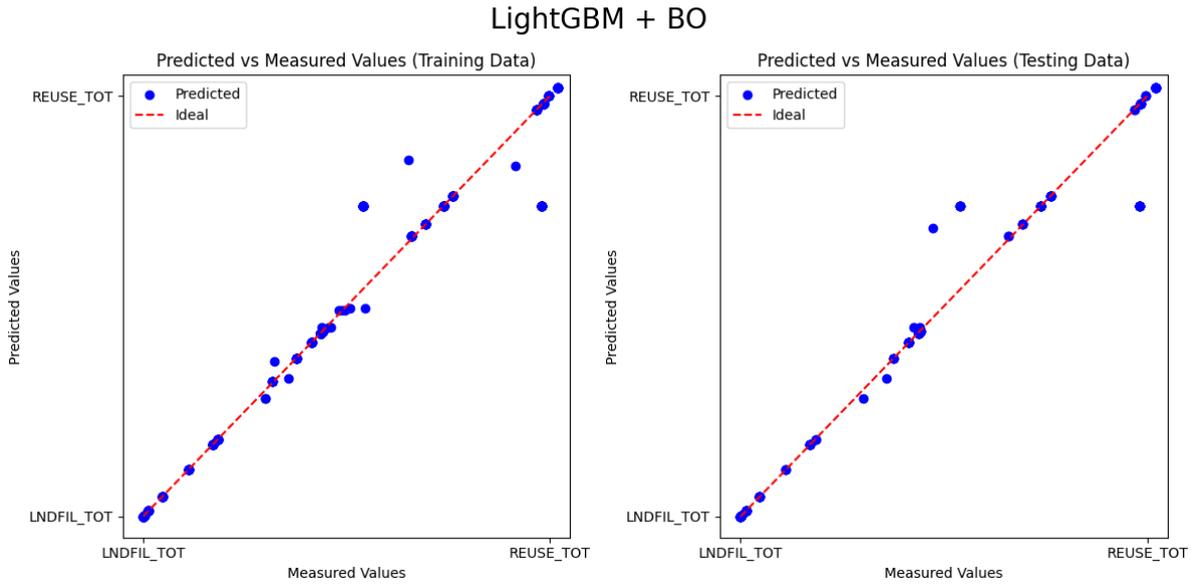

Figure 12: Actual value versus predicted value for the demolition materials – LightGBM +BO

### 3.1.1 Selection of the best ML model

As discussed in the previous section, the optimized ML models demonstrated strong predictive capabilities on the training and testing datasets. However, determining the singular overall best model requires a more holistic evaluation, as different models achieved the top performance on specific metrics. For instance, on the testing dataset, XGBoost exhibited the lowest RMSE of 40.3379, indicating the least absolute errors. DT had the lowest MAE of 3.1630, which is the smallest average error magnitude. Both DT and RF achieved the same $R^2$ of 0.9976 and NSE of 0.9821, demonstrating high explanatory power and prediction consistency.

Based on the Copeland scores, XGBoost emerged as the best-performing model with the highest score of 22 and a maximum 4 wins against all other models without any losses. KNN was ranked the lowest at 5th, with 4 losses against all competitors and a Copeland score of -26. RF took the 2nd position with a Copeland score of 9 and 3 wins. DT and LightGBM were ranked in the middle positions. Hence, the Copeland ranking identified XGBoost as the most robust ML model based on its consistent and superior performance across the diverse evaluation metrics on the testing dataset. The integrated multi-criteria assessment enabled XGBoost to be conclusively selected as the overall best predictive model for estimating demolition waste quantities

**Table 6: Result of the Copeland algorithm**

| Models | Copeland scores | Wins | Losses | Rank |
| --- | --- | --- | --- | --- |
| KNN + BO | -26 | 0 | 4 | 5 |
| DT + BO | 9 | 2 | 2 | 3 |
| RF + BO | 9 | 3 | 1 | 2 |
| XGBoost + BO | 22 | 4 | 0 | 1 |
| LightGBM + BO | -14 | 1 | 3 | 4 |

## 3.2 System Development
The system development is categorized into backend development, frontend development, and validation & verification, as discussed in the following subsections.

### 3.2.1 Backend Development
XGBoost optimized with Bayesian Optimization emerged as the best-performing model with the highest score of 22 and a maximum 4 wins against all other models without any losses. Consequently, it is pickled and cached to optimize loading times and enhance user experience, as shown in Figure 13.

```python
import streamlit as st
import requests
import json
import pickle
import streamlit.components.v1 as components
import pandas as pd
from dotenv import load_dotenv
load_dotenv()
import os

# Load machine learning model
#@st.cache
@st.cache_data
def load_model():
    with open('model.pkl', 'rb') as file:
        model = pickle.load(file)
    return model

model = load_model()
```

Figure 13: Loading of Pickled Best-Performing Model

To access the BIM data from Autodesk Forge, the application authenticates using client credentials. Also, the Forge Viewer is embedded into the application for 3D viewing and interaction with the model, as shown in Figure 14.

```python
# Function to get the access token
def get_access_token(client_id, client_secret):
    data = {
        'client_id': client_id,
        'client_secret': client_secret,
        'grant_type': 'client_credentials',
        'scope': 'data:read'
    }
    response = requests.post('https://developer.api.autodesk.com/authentication/v1/authenticate', data=data)
    if response.status_code == 200:
        return response.json()['access_token']
    else:
        st.error("Failed to get access token")
        return None

# embedding the Autodesk Forge Viewer
def create_forge_viewer_html(urn, access_token):
    return f"""
    <!DOCTYPE html>
    <html>
    <head>
        <meta charset="utf-8">
        <title>Forge Viewer</title>
        <style>
            body {{ margin: 0; }}
            #forgeViewer {{ width: 100%; height: 100%; }}
        </style>
    </head>
    <body>
        <div id="forgeViewer"></div>

        <script src="https://developer.api.autodesk.com/modelderivative/v2/viewers/7.*/viewer3D.min.js"></script>
        <script>
            var viewer;
            function launchViewer(urn) {{
                var options = {{
                    env: 'AutodeskProduction',
                    getAccessToken: function (onTokenReady) {{
                        var token = '{access_token}';
                        var timeInSeconds = 3600; // Token expiration time
                        onTokenReady(token, timeInSeconds);
                    }}
                }};
```

Figure 14: Integration of Autodesk Forge for BIM data

### 3.2.2 Frontend Development

The create_forge_viewer_html function generates HTML and JavaScript code to embed the Autodesk Forge Viewer in the Streamlit app, as shown in Figure 15.

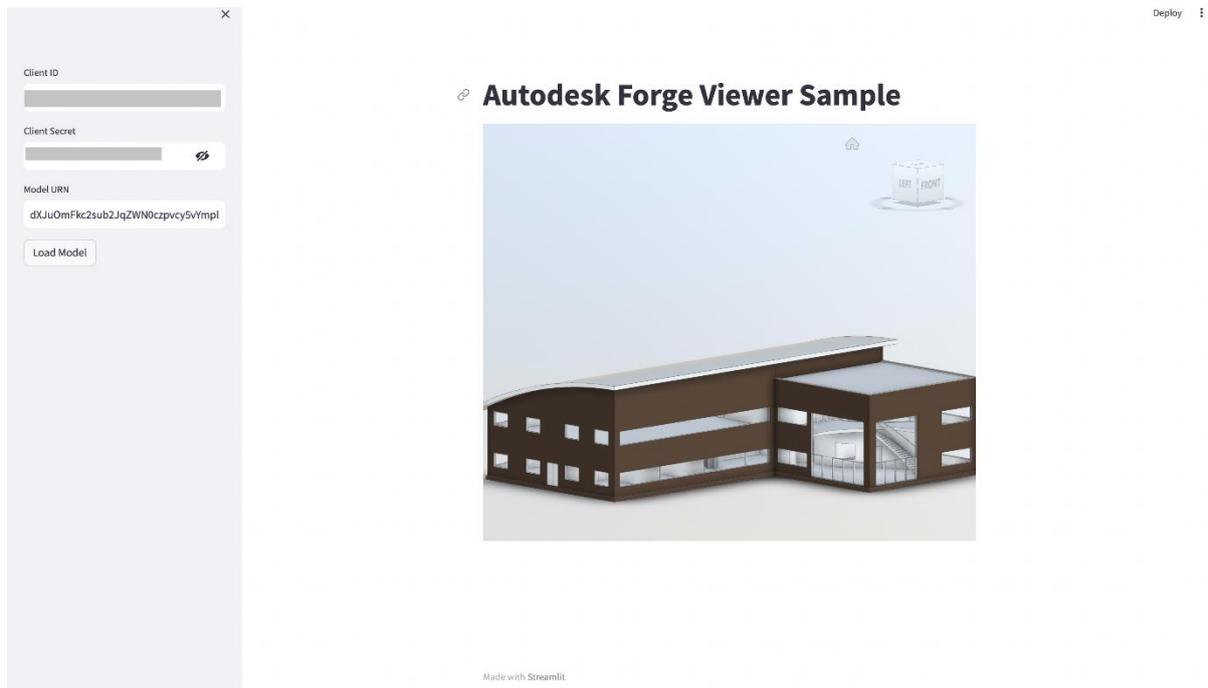

Figure 15: Autodesk Forge Viewer

The features for the prediction extracted from the BIM model were integrated into the system as shown in Figure 16.

```python
def main():
    st.title("AI-enabled BIM platform for Demolition prediction")

    # Input fields for client ID and secret
    #client_id = st.sidebar.text_input("Client ID")
    #client_secret = st.sidebar.text_input("Client Secret", type="password")
    #urn = st.sidebar.text_input("Model URN")
    client_id = os.environ.get('FORGE_CLIENT_ID')
    client_secret = os.environ.get('FORGE_CLIENT_SECRET')
    urn = os.environ.get('URN')

    # Input fields for the machine learning model
    st.sidebar.subheader("Project Inputs")
    #gfa = st.sidebar.number_input("GFA", step=1.0)
    gfa = os.environ.get('AREA')
    st.sidebar.write(f"GFA: {gfa}m²")
    #volume = st.sidebar.number_input("Volume", step=1.0)
    volume = os.environ.get('VOLUME')
    st.sidebar.write(f"Volume: {volume:.2f}m²")
    #no_of_level = st.sidebar.number_input("Number of Levels", step=1)
    no_of_level = os.environ.get('LEVEL')
    st.sidebar.write(f"Number of Levels: {no_of_level}")

    frame_types = ["Concrete", "Masonry", "Steel", "Timber"]
    usage_types = ["Agricultural", "Education", "Factory", "Hospital", "Offices", "Residential", "Retail"]

    frame_type = st.sidebar.radio("Frame Type", options=frame_types)
    usage_type = st.sidebar.radio("Usage Type", options=usage_types)
```

Figure 16: Streamlit Application Development

Lastly, the parameters are passed to the pre-trained ML model, and prediction results are displayed as shown in Figure 17.

```python
# Button to make predictions
if st.sidebar.button("Make Prediction"):
    # Prepare input data for the model
    input_data = pd.DataFrame([{
        "GFA": gfa,
        "VOLUME": volume,
        "NO_OF_LEVEL": no_of_level
    }])

    # Fill other dummy variables with 0
    #frame_types = ["Concrete", "Masonry", "Steel", "Timber"]
    #usage_types = ["Agricultural", "Education", "Factory", "Hospital", "Offices", "Residential", "Retail"]

    for ft in frame_types:
        if ft == frame_type:
            input_data[f"FRAME_TYPE_{ft}"] = 1
        else:
            input_data[f"FRAME_TYPE_{ft}"] = 0

    for ut in usage_types:
        if ut == usage_type:
            input_data[f"USAGE_TYPE_{ut}"] = 1
        else:
            input_data[f"USAGE_TYPE_{ut}"] = 0

    # Convert boolean inputs to integer
    #input_data = input_data.astype(int)

    # Make prediction
    prediction = model.predict(input_data)
    recy_tot, reuse_tot, lndfil_tot = prediction[0]

    #st.sidebar.write(input_data)

    # Display predictions
    st.sidebar.subheader("Prediction Results")
    st.sidebar.write(f"RECYCLE : {recy_tot:.2f}m³")
    st.sidebar.write(f"REUSE : {reuse_tot:.2f}m³")
    st.sidebar.write(f"LAND FILL: {lndfil_tot:.2f}m³")
```

Figure 17: Frontend Display

### 3.2.3 Validation and Verification

Verification is 'building the product/system right', and validation deals with 'building the right product/system' (Boehm, 1984). The criteria for the verification and validation are completeness, consistency, feasibility, and testability (Saka *et al.*, 2022). These were targeted via prototype development case testing with edge case and edge path users, as shown in Figure 18.

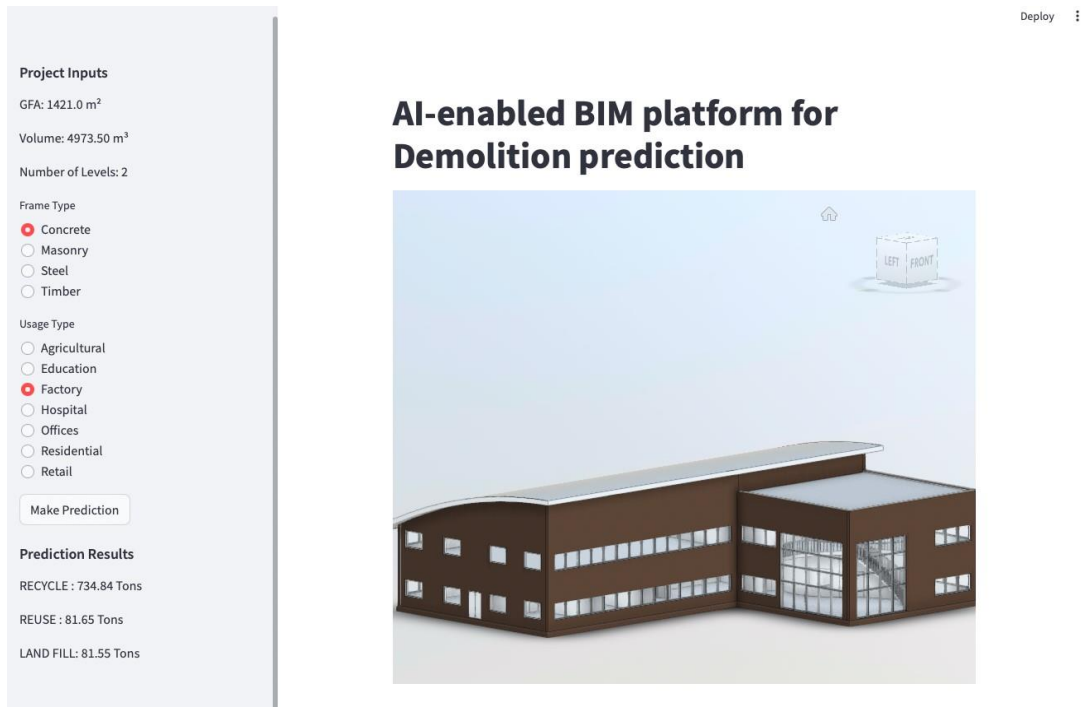

Figure 18: Case Testing

## 4.0 Discussion of Findings

The XGBoost model emerged as the best-performing model for predicting the circularity of demolition waste in construction projects. This is likely due to XGBoost being an ensemble algorithm that generally outperforms single learners. Ensemble ML techniques combine multiple learners to create a single model with superior predictive performance compared to each individual learner (Gomes *et al.*, 2018). Per Dietterich (2002), ensemble learning is often preferred because it addresses the statistical, computational, and representational limitations of single learners. In this study, the ensemble learning approach proved to be the most effective model as it reduces both bias and variance, leading to better performance than single learners. This aligns with the findings of Egwim *et al.* (2021), Yan, He*, et al.* (2022), and Yan, Yan*, et al.* (2022) on the application of XGBoost in the construction industry.

Understanding the factors that contribute to the predictions made by the selected XGBoost model is crucial for gaining insights into the underlying relationships between the features and the predicted values. To achieve this, the SHAP approach was employed, which provides a measure of feature importance and contribution to the model's predictions. The SHAP summary plot for the XGBoost model is presented in Figure 19. The plot illustrates the impact of features on the model's outputs for the three target variables – (recyclable, reusable, and landfill materials represented by Class 0, 1, and 2 respectively). The features are ranked from top to bottom in order of descending importance based on their influence on the predictions.

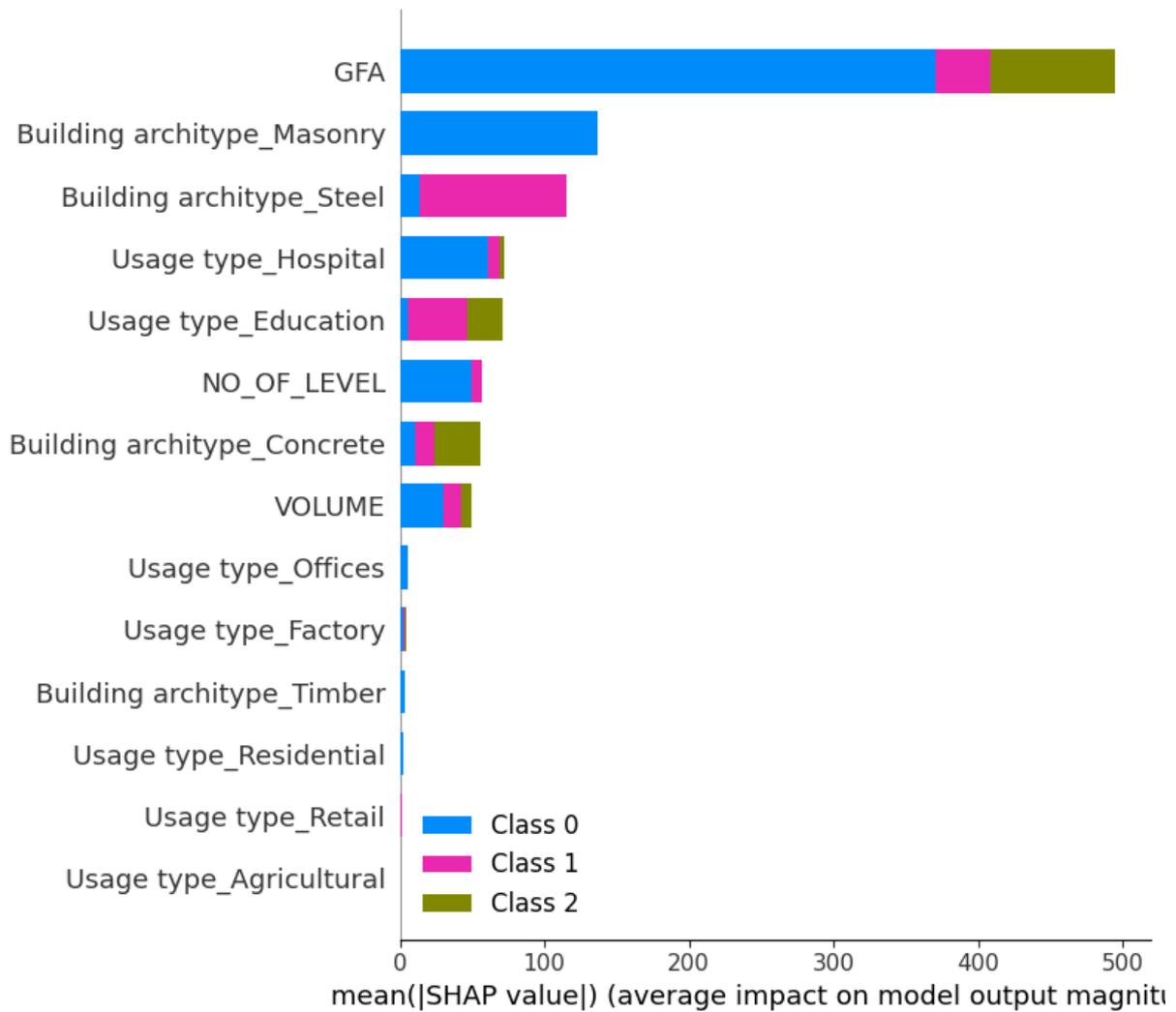

Figure 19: Contribution of each feature to the model prediction

From the plot, it can be observed that GFA has the highest impact on the model's prediction. The large positive SHAP values for GFA across all three outputs signify its strong influence in increasing the predicted waste quantities. This aligns with the insight that buildings with larger gross floor areas tend to generate greater volumes of demolition debris (Akanbi et al., 2020; Yeung et al., 2020). This is also in tandem with Soultanidis *et al.* (2023), Cha, Moon*, et al.* (2022) and Lu *et al.* (2021), that GFA is a feature for predicting construction and demolition waste. However, the impact of GFA was slightly more pronounced on the model's recyclable and landfill predictions than on reusable materials. This implies that buildings with bigger floor areas produce more quantities of materials sent for recycling or landfilling than reuse. This could be related to the design complexity of most big buildings, which have a significant impact on material reusability (Babbitt *et al.*, 2021; de Magalhaes *et al.*, 2017). As such, designers are to design for disassembly which would impact material reusability during deconstruction and demolition (Gbadamosi *et al.*, 2019).

The next feature identified was the building architectural type. Masonry and steel structural types contributed to higher predicted waste quantities in relation to other architectural styles. This broadly aligns with Menegaki and Damigos (2018) that sources of construction and demolition waste are from bricks, and steel, among others. This is understandable, as such heavy and durable constructions would logically yield more debris during demolition. Also, while the masonry archetype leads to more recyclable waste during demolition, which resonates with Ledesma *et al.* (2015), the steel archetype leans towards reusability(Sansom & Avery, 2014). Surprisingly, the timber archetype makes little contribution to predicting waste generated for reuse or landfill during building demolition. This could be related to landfills being the last alternative for timber waste due to the anaerobic decomposition of timber, which could produce methane, and recycling requiring special technologies (Kern *et al.*, 2018). On the other hand, the concrete archetype has predictive significance on recycle, reuse, and landfill options for demolition waste.

In contrast, the usage type of buildings had a relatively minor impact on the predictions, which contradicts Cha *et al.* (2023) in the South Korean context. Hospital and education-based buildings generate marginally higher reusable materials than other usage types, such as offices and factories. This could be related to the stringent requirements and need for high durability in buildings like schools and hospitals, leading to more robust materials like steel and high-grade concrete, which are more suitable for reuse. Additionally, these buildings often undergo frequent renovations to comply with updated regulations and maintain functionality, resulting in a greater quantity of high-quality reusable materials (Banias *et al.*, 2022). The least influential variables were specific usage types such as residential, retail, and agricultural - their SHAP values were near zero, indicating a negligible effect on the model's outputs.

The study also revealed the application of VM for construction and demolition waste forecasting, as shown in Table 7. The current study outperformed previous similar studies with better model predictive performance. This could be because of the dataset and algorithm employed in this study. 2280 datapoints is significantly higher than the dataset that has been leveraged in the extant studies due to the challenges of obtaining a reliable demolition dataset. Although this study employed similar algorithms to extant studies, an ensemble algorithm outperforms these algorithms. Interestingly, this study has better predictive performance ($R^2$ of 0.9977 and MAE of 3.7874) than deep learning models (Average $R^2$ of 0.97 and MAE between 17.93 -19.04) which employed a similar approach and dataset (Akanbi *et al.*, 2020). Most importantly, the current study did not lump the waste as a single output but provided circularity insights for reuse, recycling, and landfill from a single model. Lastly, compared to similar previous studies, this study developed and integrated the ML model with BIM model for data access for practical application.

Table 7: Comparison with Similar Studies

| Source | Waste type | Optimal Model | Dataset | Performance |
|---|---|---|---|---|
| Parisi Kern *et al.* (2015) | Construction | Multiple Linear Regression | 18 buildings | $R^2$ – 0.694 |
| Teixeira *et al.* (2020) | Construction | Multiple Linear Regression | 18 buildings | $R^2$ – 0.81 |
| Cha *et al.* (2020) | Demolition | Random Forest | 784 building | $R^2$ – 0.615 |
| Nagalli (2021) | Construction | ANN | 330 buildings | $R^2$ – 0.96 |
| Cha, Choi, *et al.* (2022) | Demolition | Hybrid CATPCA-SVMR | 784 building | $R^2$ – 0.594 |
| Cha *et al.* (2023) | Demolition | PCA - KNN | 160 buildings | $R^2$ – 0.897 |
| This study | Demolition | XGBoost | 2280 buildings | $R^2$ – 0.9977 |

**5.0 Conclusion**

The construction industry is the largest source of waste generation globally because it is the largest natural resource consumer. These wastes have detrimental economic, social, and environmental impacts on society. Effective management of this waste depends on properly quantifying the waste to be generated. Despite demolition waste accounting for the large part of the waste in the sector, extant studies on waste quantification have focused on construction waste. This is partly due to the challenges of obtaining demolition dataset and the lack of sorting facilities on most construction sites, resulting in bulk transfer to nearby Waste Transfer Stations (WTS). Due to this, the few extant studies on demolition waste have focused on regional level predictions, and studies at project level rely on small datasets with no circularity insights or support for practical application.

Consequently, this study modelled circularity prediction of demolition waste at the project level with the view of developing a practical tool to support pre-demolition auditing. Variable modelling approach of waste quantification was leveraged using machine learning with 2280 demolition dataset. The study revealed that the best-performing model is XGBoost with Bayesian Optimization – an ensemble learning algorithm. Unlike previous studies focusing on single-output models, this study developed a multi-output prediction. This model estimates the total quantity of demolition waste and categorizes it into recyclable, reusable, and landfill materials. This granularity aids in better planning and management of demolition waste. A practical, easy-to-use system was developed by integrating the ML and BIM models via Autodesk Forge for pre-demolition auditing. The study highlighted the high predictive performance of machine learning models over extant models, including deep learning models. This underscores the notion that simple models are often sufficient in explaining relationships and dynamics between variables in small data sizes (less than tens of thousands of data points). It highlighted that the features with predictive importance on waste generated are GFA, building archetype, and the number of levels with varying levels of significance on reuse, recycle, and landfill. The findings implied the need to design for disassembly to increase the quantity of waste that can be reused during demolition. The integrated BIM-ML system saves time compared to the manual approach of demolition quantification via site visits' direct and indirect measurement. It would benefit stakeholders in forecasting demolition waste, which would contribute to effective waste management.

The limitation of the current study serves as fertile areas for further research. As such, further study could integrate classification system accumulation (CSA) to forecast reuse, recycle, and landfill quantities for each material in the building during demolition. Although the employed dataset is extensive and the largest in extant studies, it is limited to the UK and does not capture other important features such as local regulation, demolition technique, building age, and historical maintenance, which could affect the generalizability of the study. Also, integrating dynamic features like sensor data and real-time monitoring could enhance model's adaptability and accuracy. In addition, practical implementation of the more robust developed system could be conducted, and performance evaluated by comparing forecasted waste with actual generated demolition waste. Lastly, user feedback could be gathered on the usability of the system to guide further improvements and refinements.

**References.**

Ajayi, S. O., & Oyedele, L. O. (2017). Policy imperatives for diverting construction waste from landfill: Experts' recommendations for UK policy expansion. *Journal of Cleaner Production*, *147*, 57-65. https://doi.org/10.1016/j.jclepro.2017.01.075

Abyani, M., Bahaari, M. R., Zarrin, M., & Nasseri, M. (2022). Predicting failure pressure of the corroded offshore pipelines using an efficient finite element based algorithm and machine learning techniques. Ocean Engineering, 254(November 2021), 111382. https://doi.org/10.1016/j.oceaneng.2022.111382

Akanbi, L. A., Oyedele, A. O., Oyedele, L. O., & Salami, R. O. (2020). Deep learning model for Demolition Waste Prediction in a circular economy. *Journal of Cleaner Production*, *274*. https://doi.org/10.1016/j.jclepro.2020.122843

Akanbi, L. A., Oyedele, L. O., Akinade, O. O., Ajayi, A. O., Davila Delgado, M., Bilal, M., & Bello, S. A. (2018). Salvaging building materials in a circular economy: A BIM-based whole-life performance estimator. *Resources, Conservation and Recycling*, *129*, 175-186. https://doi.org/10.1016/j.resconrec.2017.10.026

Babbitt, C. W., Althaf, S., Cruz Rios, F., Bilec, M. M., & Graedel, T. E. (2021). The role of design in circular economy solutions for critical materials. *One Earth*, *4*(3), 353-362. https://doi.org/10.1016/j.oneear.2021.02.014

Banias, G. F., Karkanias, C., Batsioula, M., Melas, L. D., Malamakis, A. E., Geroliolios, D., Skoutida, S., & Spiliotis, X. (2022). Environmental Assessment of Alternative Strategies for the Management of Construction and Demolition Waste: A Life Cycle Approach. *Sustainability*, *14*(15). https://doi.org/10.3390/su14159674

Bernardo, M., Gomes, M. C., & de Brito, J. (2016). Demolition waste generation for development of a regional management chain model. *Waste Manag*, *49*, 156-169. https://doi.org/10.1016/j.wasman.2015.12.027

Bello, I. T., Taiwo, R., Esan, O. C., Adegoke, A. H., Ijaola, A., Li, Z., Zhao, S., Chen, W., Shao, Z., & Ni, M. (2023). AI-enabled Materials Discovery for Advanced Ceramic Electrochemical Cells. Energy and AI.

Ben Seghier, M. E. A., Höche, D., & Zheludkevich, M. (2022). Prediction of the internal corrosion rate for oil and gas pipeline: Implementation of ensemble learning techniques. Journal of Natural Gas Science and Engineering, 99(January). https://doi.org/10.1016/j.jngse.2022.104425

Bilal, M., Khan, K. I. A., Thaheem, M. J., & Nasir, A. R. (2020). Current state and barriers to the circular economy in the building sector: Towards a mitigation framework. *Journal of Cleaner Production*, *276*. https://doi.org/10.1016/j.jclepro.2020.123250

Boehm, B. W. (1984). Verifying and Validating Software Requirements and Design Specifications. *IEEE Software*, *1*(1), pp. 75-88. https://doi.org/10.1109/MS.1984.233702

Breiman, L. (2001). Random Forests. Machine Learning, 45(1), 5–32.

Breiman, L., Friedman, J. H., Olshen, R. A., & Stone, C. (1984). Classification and Regression Trees. Biometrics, 40(3).

Cha, G.-W., Moon, H. J., & Kim, Y.-C. (2022). A hybrid machine-learning model for predicting the waste generation rate of building demolition projects. *Journal of Cleaner Production*, *375*. https://doi.org/10.1016/j.jclepro.2022.134096

Cha, G. W., Choi, S. H., Hong, W. H., & Park, C. W. (2022). Development of Machine Learning Model for Prediction of Demolition Waste Generation Rate of Buildings in Redevelopment Areas. *Int J Environ Res Public Health*, *20*(1). https://doi.org/10.3390/ijerph20010107

Cha, G. W., Choi, S. H., Hong, W. H., & Park, C. W. (2023). Developing a Prediction Model of Demolition-Waste Generation-Rate via Principal Component Analysis. *Int J Environ Res Public Health*, *20*(4). https://doi.org/10.3390/ijerph20043159

Cha, G. W., Moon, H. J., Kim, Y. M., Hong, W. H., Hwang, J. H., Park, W. J., & Kim, Y. C. (2020). Development of a Prediction Model for Demolition Waste Generation Using a Random Forest

Algorithm Based on Small DataSets. *Int J Environ Res Public Health*, *17*(19). https://doi.org/10.3390/ijerph17196997

Chen, T., & Guestrin, C. (2016). XGBoost: A scalable tree boosting system. Proceedings of the ACM SIGKDD International Conference on Knowledge Discovery and Data Mining, 13-17-Augu, 785–794. https://doi.org/10.1145/2939672.2939785

Cheng, J. C., & Ma, L. Y. (2013). A BIM-based system for demolition and renovation waste estimation and planning. *Waste Manag*, *33*(6), 1539-1551. https://doi.org/10.1016/j.wasman.2013.01.001

Coskuner, G., Jassim, M. S., Zontul, M., & Karateke, S. (2021). Application of artificial intelligence neural network modeling to predict the generation of domestic, commercial and construction wastes. *Waste Manag Res*, *39*(3), 499-507. https://doi.org/10.1177/0734242X20935181

Daoud, E. Al. (2019). Comparison between XGBoost, LightGBM and CatBoost Using a Home Credit Dataset. International Journal of Computer and Information Engineering, 13(1), 6–10.

de Guzman Baez, A., Villoria Saez, P., del Rio Merino, M., & Garcia Navarro, J. (2012). Methodology for quantification of waste generated in Spanish railway construction works. *Waste Manag*, *32*(5), 920-924. https://doi.org/10.1016/j.wasman.2012.01.007

de Magalhaes, R. F., Danilevicz, A. M. F., & Saurin, T. A. (2017). Reducing construction waste: A study of urban infrastructure projects. *Waste Manag*, *67*, 265-277. https://doi.org/10.1016/j.wasman.2017.05.025

Dietterich, G. T. (2002). Ensemble Learning. In M. A. Arbib (Ed.), *The Handbook of Brain Theory and Neural Network* (Second ed.). MIT Press.

Egwim, C. N., Alaka, H., Toriola-Coker, L. O., Balogun, H., & Sunmola, F. (2021). Applied artificial intelligence for predicting construction projects delay. *Machine Learning with Applications*, *6*. https://doi.org/10.1016/j.mlwa.2021.100166

Furxhi, I., Murphy, F., Mullins, M., & Poland, C. A. (2019). Machine learning prediction of nanoparticle in vitro toxicity: A comparative study of classifiers and ensemble-classifiers using the Copeland Index. Toxicology Letters, 312(May), 157–166. https://doi.org/10.1016/j.toxlet.2019.05.016

Gbadamosi, A.-Q., Mahamadu, A.-M., Oyedele, L. O., Akinade, O. O., Manu, P., Mahdjoubi, L., & Aigbavboa, C. (2019). Offsite construction: Developing a BIM-Based optimizer for assembly. *Journal of Cleaner Production*, *215*, 1180-1190. https://doi.org/10.1016/j.jclepro.2019.01.113

Gomes, H. M., Barddal, J. P., Enembreck, F., & Bifet, A. (2018). A Survey on Ensemble Learning for Data Stream Classification. *ACM Computing Surveys*, *50*(2), 1-36. https://doi.org/10.1145/3054925

Islam, R., Nazifa, T. H., Yuniarto, A., Shanawaz Uddin, A. S. M., Salmiati, S., & Shahid, S. (2019). An empirical study of construction and demolition waste generation and implication of recycling. *Waste Manag*, *95*, 10-21. https://doi.org/10.1016/j.wasman.2019.05.049

Kamma, R. C., & Jha, K. N. (2022). Quantifying Building Construction and Demolition Waste Using Permit Data. *Journal of Construction Engineering and Management*, *148*(9). https://doi.org/10.1061/(asce)co.1943-7862.0002357

Kartam, N., Al-Mutairi, N., Al-Ghusain, I., & Al-Humoud, J. (2004). Environmental management of construction and demolition waste in Kuwait. *Waste Manag*, *24*(10), 1049-1059. https://doi.org/10.1016/j.wasman.2004.06.003

Ke, G., Meng, Q., Finley, T., Wang, T., Chen, W., Ma, W., Ye, Q., & Liu, T. Y. (2017). LightGBM: A highly efficient gradient boosting decision tree. Advances in Neural Information Processing Systems, 2017-Decem(Nips), 3147–3155

Kern, A. P., Amor, L. V., Angulo, S. C., & Montelongo, A. (2018). Factors influencing temporary wood waste generation in high-rise building construction. *Waste Manag*, *78*, 446-455. https://doi.org/10.1016/j.wasman.2018.05.057

Lam, P. T. I., Yu, A. T. W., Wu, Z., & Poon, C. S. (2019). Methodology for upstream estimation of construction waste for new building projects. *Journal of Cleaner Production*, *230*, 1003-1012. https://doi.org/10.1016/j.jclepro.2019.04.183

Ledesma, E. F., Jiménez, J. R., Ayuso, J., Fernández, J. M., & de Brito, J. (2015). Maximum feasible use of recycled sand from construction and demolition waste for eco-mortar production – Part-I: ceramic masonry waste. *Journal of Cleaner Production*, *87*, 692-706. https://doi.org/10.1016/j.jclepro.2014.10.084

Li, Y., & Zhang, X. (2013). Web-based construction waste estimation system for building construction projects. *Automation in Construction*, *35*, 142-156. https://doi.org/10.1016/j.autcon.2013.05.002

Lu, W., Lou, J., Webster, C., Xue, F., Bao, Z., & Chi, B. (2021). Estimating construction waste generation in the Greater Bay Area, China using machine learning. *Waste Manag*, *134*, 78-88. https://doi.org/10.1016/j.wasman.2021.08.012

Lundberg, S., & Lee, S.-I. (2017). A Unified Approach to Interpreting Model Predictions. 31st Conference on Neural Information Processing Systems, 4, 552–564. https://doi.org/10.1016/j.ophtha.2018.11.016

Malia, M., de Brito, J., Pinheiro, M. D., & Bravo, M. (2013). Construction and demolition waste indicators. *Waste Manag Res*, *31*(3), 241-255. https://doi.org/10.1177/0734242X12471707

Menegaki, M., & Damigos, D. (2018). A review on current situation and challenges of construction and demolition waste management. *Current Opinion in Green and Sustainable Chemistry*, *13*, 8-15. https://doi.org/10.1016/j.cogsc.2018.02.010

Nagalli, A. (2021). Estimation of construction waste generation using machine learning. *Proceedings of the Institution of Civil Engineers - Waste and Resource Management*, *174*(1), 22-31. https://doi.org/10.1680/jwarm.20.00019

Oluleye, B. I., Chan, D. W. M., Saka, A. B., & Olawumi, T. O. (2022). Circular economy research on building construction and demolition waste: A review of current trends and future research directions. *Journal of Cleaner Production*, *357*. https://doi.org/10.1016/j.jclepro.2022.131927

Osmani, M. (2011). Construction Waste. In *Waste* (pp. 207-218). https://doi.org/10.1016/b978-0-12-381475-3.10015-4

Parisi Kern, A., Ferreira Dias, M., Piva Kulakowski, M., & Paulo Gomes, L. (2015). Waste generated in high-rise buildings construction: a quantification model based on statistical multiple regression. *Waste Manag*, *39*, 35-44. https://doi.org/10.1016/j.wasman.2015.01.043

Patrick, E. A., & Fischer, F. P. (1970). A generalized k-nearest neighbor rule. Information and Control, 16(2), 128–152. https://doi.org/10.1016/S0019-9958(70)90081-1

Pheng, L. S., & Hou, L. S. (2019). The Economy and the Construction Industry. In *Construction Quality and the Economy* (pp. 21-54). https://doi.org/10.1007/978-981-13-5847-0_2

Saka, A. B., Chan, D. W. M., & Wuni, I. Y. (2022). Knowledge-based decision support for BIM adoption by small and medium-sized enterprises in developing economies. *Automation in Construction*, *141*. https://doi.org/10.1016/j.autcon.2022.104407

Saka, A. B., Olaore, F., & Olawumi, T. O. (2019). Post-contract material management and waste minimization: an analysis of the roles of quantity surveyors. *Journal of Engineering, Design and Technology*, *17*(4), 793-807. https://doi.org/10.1108/JEDT-10-2018-0193

Sansom, M., & Avery, N. (2014). Briefing: Reuse and recycling rates of UK steel demolition arisings. *Proceedings of the Institution of Civil Engineers - Engineering Sustainability*, *167*(3), 89-94. https://doi.org/10.1680/ensu.13.00026

Samet, H. (2008). K-nearest neighbor finding using MaxNearestDist. IEEE Transactions on Pattern Analysis and Machine Intelligence, 30(2), 243–252. https://doi.org/10.1109/TPAMI.2007.1182

Sun, D., Xu, J., Wen, H., & Wang, D. (2021). Assessment of landslide susceptibility mapping based on Bayesian hyperparameter optimization: A comparison between logistic regression and random forest. Engineering Geology, 281(May 2020), 105972. https://doi.org/10.1016/j.enggeo.2020.105972

Shehadeh, A., Alshboul, O., Al Mamlook, R. E., & Hamedat, O. (2021). Machine learning models for predicting the residual value of heavy construction equipment: An evaluation of modified decision tree, LightGBM, and XGBoost regression. *Automation in Construction*, *129*. https://doi.org/10.1016/j.autcon.2021.103827

Shmueli, G. (2010). To Explain or to Predict? *Statistical Science*, *25*(3). https://doi.org/10.1214/10-sts330

Soultanidis, V., & Voudrias, E. A. (2023). Modelling of demolition waste generation: Application to Greek residential buildings. *Waste Manag Res*, *41*(9), 1469-1479. https://doi.org/10.1177/0734242X231155818

Taiwo, R., Ben Seghier, M. E. A., & Zayed, T. (2023a). Predicting Wall Thickness Loss in Water Pipes Using Machine Learning Techniques. 2nd Conference of the European Association on Quality Control of Bridges and Structures - EUROSTRUCT2023. 10.1002/cepa.2075

Taiwo, R., Ben Seghier, M. E. A., & Zayed, T. (2023b). Towards sustainable water infrastructure : The state-of-the-art for modeling the failure probability of water pipes. Water Resources Research.

Taiwo, R., Hussein, M., & Zayed, T. (2022). An Integrated Approach of Simulation and Regression Analysis for Assessing Productivity in Modular Integrated Construction Projects. Buildings, 12. https://doi.org/https://doi.org/10.3390/ buildings12112018

Taiwo, R., Yussif, A., El, M., Ben, A., & Zayed, T. (2024). Explainable ensemble models for predicting wall thickness loss of water pipes Area Under the Curve. Ain Shams Engineering Journal, August 2023, 102630. https://doi.org/10.1016/j.asej.2024.102630

Taiwo, R., Zayed, T., & Ben Seghier, M. E. A. (2024). Integrated intelligent models for predicting water pipe failure probability. Alexandria Engineering Journal, 86, 243–257. https://doi.org/10.1016/j.aej.2023.11.047

Teixeira, E. C., Gonzalez, M. A. S., Heineck, L. F. M., Kern, A. P., & Bueno, G. M. (2020). Modelling waste generated during construction of buildings using regression analysis. *Waste Manag Res*, *38*(8), 857-867. https://doi.org/10.1177/0734242X19893012

Wu, Z., Yu, A. T., Shen, L., & Liu, G. (2014). Quantifying construction and demolition waste: an analytical review. *Waste Manag*, *34*(9), 1683-1692. https://doi.org/10.1016/j.wasman.2014.05.010

Wang, J., & Ashuri, B. (2016). Predicting ENR's Construction Cost Index Using the Modified K Nearest Neighbors (KNN) Algorithm. PROCEEDINGS Construction Research Congress 2016, 100, 2039–2049. https://doi.org/10.1061/9780784479827.203

Yan, H., He, Z., Gao, C., Xie, M., Sheng, H., & Chen, H. (2022). Investment estimation of prefabricated concrete buildings based on XGBoost machine learning algorithm. *Advanced Engineering Informatics*, *54*. https://doi.org/10.1016/j.aei.2022.101789

Yan, H., Yan, K., & Ji, G. (2022). Optimization and prediction in the early design stage of office buildings using genetic and XGBoost algorithms. *Building and Environment*, *218*. https://doi.org/10.1016/j.buildenv.2022.109081

Yeung, H. C., Ridwan, T., Tariq, S., & Zayed, T. (2020). BEAM Plus implementation in Hong Kong: assessment of challenges and policies. International Journal of Construction Management. https://doi.org/10.1080/15623599.2020.1827692

Yussif, A. M., Sadeghi, H., & Zayed, T. (2023). Application of Machine Learning for Leak Localization in Water Supply Networks. Buildings, 13(4), 1–21. https://doi.org/10.3390/buildings13040849

Yu, B., Wang, J., Li, J., Zhang, J., Lai, Y., & Xu, X. (2019). Prediction of large-scale demolition waste generation during urban renewal: A hybrid trilogy method. *Waste Manag*, *89*, 1-9. https://doi.org/10.1016/j.wasman.2019.03.063

Yuan, H., & Shen, L. (2011). Trend of the research on construction and demolition waste management. *Waste Manag*, *31*(4), 670-679. https://doi.org/10.1016/j.wasman.2010.10.030

Victoria, A. H., & Maragatham, G. (2021). Automatic tuning of hyperparameters using Bayesian optimization. Evolving Systems, 12(1), 217–223. https://doi.org/10.1007/s12530-020-09345-2